\documentclass[conference]{IEEEtran}
\IEEEoverridecommandlockouts
\usepackage{cite}
\usepackage{amsmath,amssymb,amsfonts}
\usepackage{algorithmic}
\usepackage{graphicx}
\usepackage{textcomp}
\usepackage{xcolor}
\usepackage{hyperref}
\begin{document}

\title{Masked Face Inpainting Through Residual Attention UNet\\

}

\author{\IEEEauthorblockN{Md Imran Hosen}
\IEEEauthorblockA{\textit{Department of Computer Engineering} \\
\textit{Bahcesehir University, Istanbul, Turkey}\\
imranjucse21@gmail.com}
\and
\IEEEauthorblockN{Md Baharul Islam}
\IEEEauthorblockA{\textit{Bahcesehir University, Istanbul, Turkey} \\
\textit{American University of Malta}\\
bislam.eng@gmail.com }
}

\maketitle

\IEEEpubidadjcol

\begin{abstract}
Realistic image restoration with high texture areas such as removing face masks is challenging. The state-of-the-art deep learning-based methods fail to guarantee high-fidelity, cause training instability due to vanishing gradient problems (e.g., weights are updated slightly in initial layers) and spatial information loss. They also depend on intermediary stage such as segmentation meaning require external mask.  This paper proposes a blind mask face inpainting method using residual attention UNet to remove the face mask and restore the face with fine details while minimizing the gap with the ground truth face structure. A residual block feeds info to the next layer and directly into the layers about two hops away to solve the gradient vanishing problem. Besides, the attention unit helps the model focus on the relevant mask region, reducing resources and making the model faster. Extensive experiments on the publicly available CelebA dataset show the feasibility and robustness of our proposed model. Code is available at \url{https://github.com/mdhosen/Mask-Face-Inpainting-Using-Residual-Attention-Unet}
\end{abstract}


\section{Introduction}
Image inpainting (also known as image completion) is the task of removing unwanted content that has many applications such as face mask removing or selected object replacement, damaged paintings restoration \cite{deng2020image}. To minimize the spread of infectious disease (e.g., COVID-19), wearing face masks in public areas makes the security and identification system vulnerable. Thus, restoring covered face is an interesting problem with significant real-world applications.

Many deep learning-based image inpainting methods \cite{yu2018generative,lahiri2019faster, zheng2019pluralistic,li2020interactive, din2020novel, zhao2021comodgan,yang2020generative, wang2020vcnet, jiang2022mask} have achieved satisfactory performance for removing an unwanted object from the image. Yu et al. \cite{yu2018generative} proposed a contextual attention-based image inpainting model with two generators and two discriminators that get the relevant data from far-off spatial areas for reproducing the missing pixels. The coarse reconstruction network estimates the missing areas roughly while the contextual attention is implemented to the second refinement network. The local and global discriminators capture the generated pixel's texture details. However, this method showed moderate performance for small-scale and free-form shapes and failed to restore the large area. A similar limitation has occurred in \cite{lahiri2019faster,yang2020generative, wang2020vcnet}.

\begin{figure}[t]
\centering
\begin{minipage}{1\textwidth}
\hspace{1cm} Input \hspace{2cm} GT \hspace{2cm} Output
\end{minipage}
\includegraphics[width=0.14\textwidth]{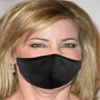}
\includegraphics[width=0.14\textwidth]{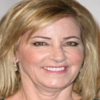}
\includegraphics[width=0.14\textwidth]{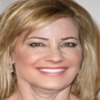}
\caption{\label{output}An example of mask face inpainting using our method.}
\end{figure}

A probabilistic framework with two parallel pipelines (e.g., reconstructive and generative path) proposed in \cite{zheng2019pluralistic,deng2020image} that performs multiple image restoration. Although Zheng et al. \cite{zheng2019pluralistic} obtained realistic results for large-scale free form shape that suffers intensive computation and training instability \cite{jiang2022mask}. Din et al.,\cite{din2020novel} dealt with this challenge by applying a two-stage Generative Adversarial Network (GAN), including binary segmentation and image restoration that is faster. However, it faces a color inconsistency problem to restore the original structure  and depend on intermediary stage such as segmentation. These methods may not be usefull for removing mask face area without distortion of the face structure. Thus, an efficient, faster, and highly accurate blind inpainting method is required to remove the large mask face area, avoid the vanishing gradient problem and does not depend on intermediate stage such as segmentation, facial landmark prediction \cite{yang2020generative} . To address these issues, we propose a mask face inpainting model based on residual attention UNet that can automatically remove and restore the face mask area with a realistic view and fine details. An example of the mask face completion has been demonstrated in Fig. \ref{output} using our method. The significant contributions are summarized below. 

\begin{figure*}[htb]
\centerline{\includegraphics[width=1\textwidth]{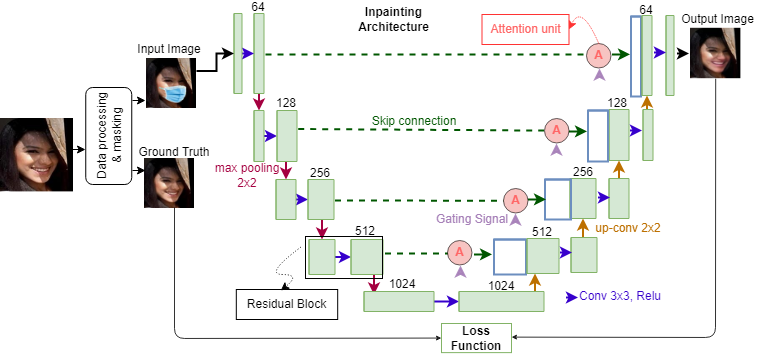}}
\caption{The architecture of the proposed framework with residual and attention-based UNet for mask face inpainting is shown.}
\label{methodologhy}
\end{figure*}

\begin{itemize}
    \item A novel mask face inpainting model has been proposed to remove the mask and restore the face structure automatically with fine details.
    
    \item To address the vanishing gradient problem and make the train stable, we added a residual block that passes the information into the layer in two hops away.
    
    \item We use an attention unit that focuses on the relevant area, reduces resources, and makes the model faster.
    \item Our technique is blind image inpainting. It does not depend on intermediate stages such as mask segmentation.
    \item An extensive experiment and ablation study was performed to validate the proposed method for the large-scale CelebA \cite{liu2015faceattributes} dataset.
\end{itemize} 

\section{Proposed Method}
\label{sec:methd}

The architecture of the proposed method has been demonstrated in Fig. \ref{methodologhy}. We use UNet \cite{ronneberger2015u} with residual block and attention unit. The encoder parts consist of 5 convolution stacked. Two consecutive convolution operations with size $3\times3$ have been performed in each convolution stacked, followed by an activation function ReLU and generate feature map ($F$). A $2\times2$ max-pooling layer is performed with stride 2 after $F$ is concatenated with input through a $1\times1$ residual path. The decoder part in the UNet is identical to the encoder part, except the decoder uses $2\times2$ up-sampling instead of $2\times2$ down-sampling. In the end, the sigmoid activation function is used to predict whether the pixel belongs to the point. In following subsections, we discuss both units in details.  

\subsection{Residual Block}

\begin{figure}[h!]
\centerline{\includegraphics[angle=-90, width=0.22\textwidth]{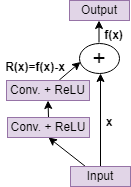}}
\caption{Residual block that sends the information to the next layer and directly into the layers 2 hops away.}
\label{residual}
\end{figure}

We use the residual block \cite{he2016deep} to overcome the problems of deep learning-based networks, such as the vanishing gradient problem, meaning weights are updated ineffectively. Since these wights are multiplied; if propagating continues weights become insignificant when it reaches the first layer. Therefore, the increment in weights gets very small, the model is not changing, which happens with intense neural networks. Adding more layers makes the probability of making the model overfitting. Fig. \ref{residual}  shows the residual block. Instead of function $f(x)$ , model learns from $f(x)$ -$x$ in residual block. We add $f(x)$-$x$ to the output. Each layer feeds into the next and directly into the layers about 2 hops away in the residual block. Forward propagation is faster through the residual connection because of the shortcut layer.

After two consecutive convolutions, we get new feature map $F( H\times W\times k^2 C)$; here $H$, $W$ and $C$ represent height, width and channel respectively and  $k$ indicates the convolution kernel. Next, F is combined with I ( input image) on the channel dimension and generate a feature map $M_f( H\times W\times (k^2+1) C)$. After that $1\times1$ convolution operation is performed to maintain the channel dimension. 

\begin{figure}[htb]
\centerline{\includegraphics[width=0.45\textwidth]{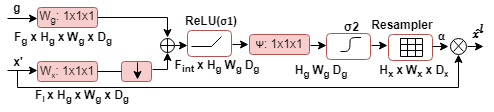}}
\caption{Attention unit activates the relevant area in training. }
\label{attention}
\end{figure}

\begin{figure*}[htb]
\centering
\begin{minipage}{1\textwidth}
\hspace{1.3cm} Input \hspace{1.9cm} GT \hspace{1.9cm} Output \hspace{2.1cm} Input \hspace{1.9cm} GT \hspace{2.1cm} Output
\end{minipage}
\includegraphics[width=0.15\textwidth]{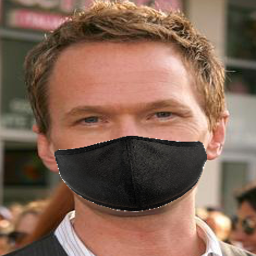}
\includegraphics[width=0.15\textwidth]{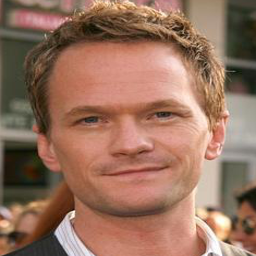}
\includegraphics[width=0.15\textwidth]{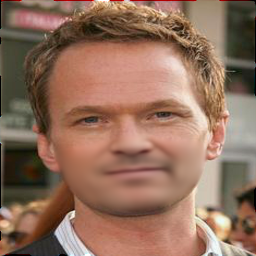}
\includegraphics[width=0.15\textwidth]{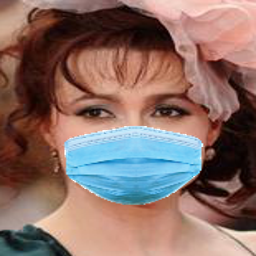}
\includegraphics[width=0.15\textwidth]{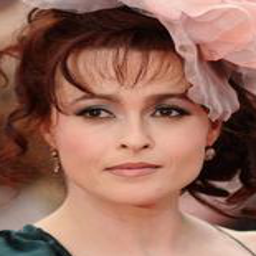}
\includegraphics[width=0.15\textwidth]{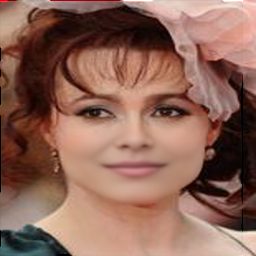}

\includegraphics[width=0.15\textwidth]{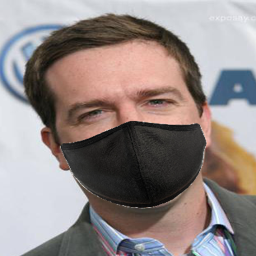}
\includegraphics[width=0.15\textwidth]{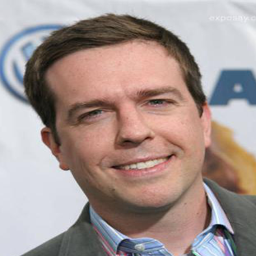}
\includegraphics[width=0.15\textwidth]{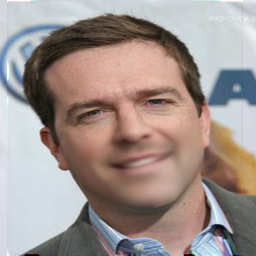}
\includegraphics[width=0.15\textwidth]{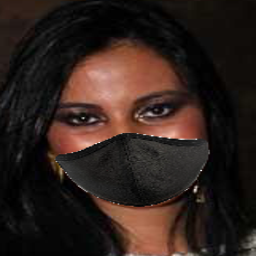}
\includegraphics[width=0.15\textwidth]{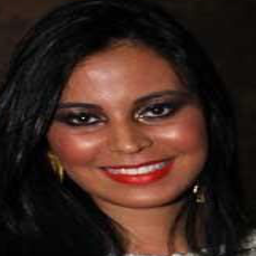}
\includegraphics[width=0.15\textwidth]{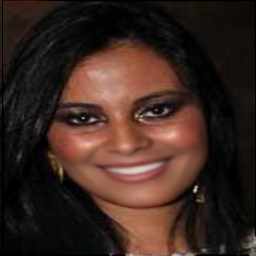}

\includegraphics[width=0.15\textwidth]{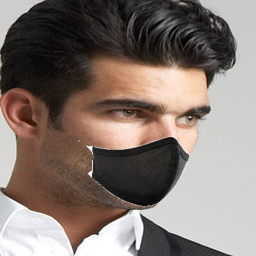}
\includegraphics[width=0.15\textwidth]{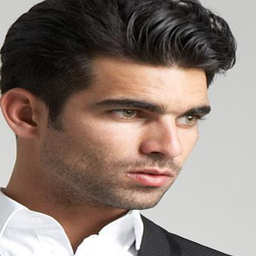}
\includegraphics[width=0.15\textwidth]{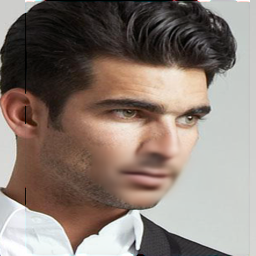}
\includegraphics[width=0.15\textwidth]{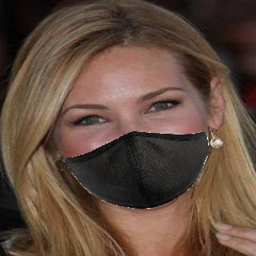}
\includegraphics[width=0.15\textwidth]{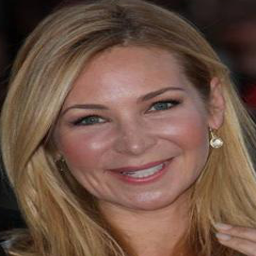}
\includegraphics[width=0.15\textwidth]{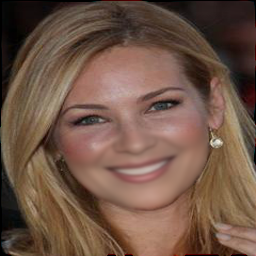}

\includegraphics[width=0.15\textwidth]{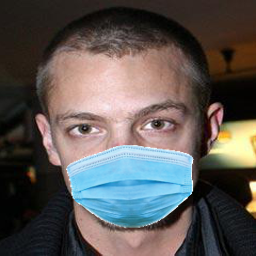}
\includegraphics[width=0.15\textwidth]{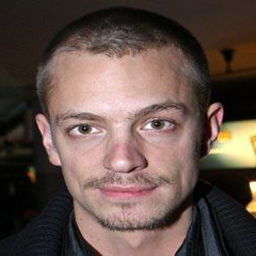}
\includegraphics[width=0.15\textwidth]{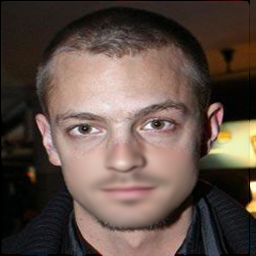}
\includegraphics[width=0.15\textwidth]{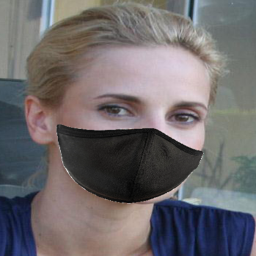}
\includegraphics[width=0.15\textwidth]{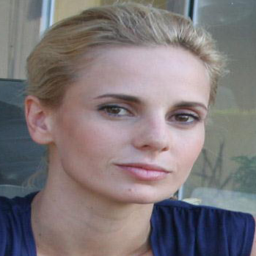}
\includegraphics[width=0.15\textwidth]{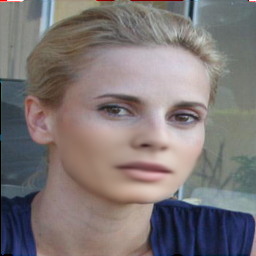}

\caption{\label{Quality} \small Face image inpainting results on CelebA dataset by our approach.}
\end{figure*}

\begin{figure*}[htb]
\centering
\begin{minipage}{1\textwidth}
\hspace{1.6cm} Input \hspace{1.7cm} GT \hspace{1.7cm} Yu et al.\cite{yu2018generative} \hspace{0.6cm} Zheng et al. \cite{zheng2019pluralistic} \hspace{0.5cm} Din et al. \cite{din2020novel} \hspace{1.7cm} Ours
\end{minipage}
\includegraphics[width=0.15\textwidth]{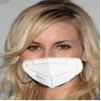}
\includegraphics[width=0.15\textwidth]{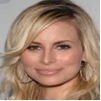}
\includegraphics[width=0.15\textwidth]{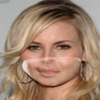}
\includegraphics[width=0.15\textwidth]{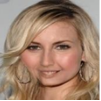}
\includegraphics[width=0.15\textwidth]{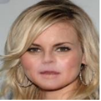}
\includegraphics[width=0.15\textwidth]{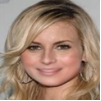}

\includegraphics[width=0.15\textwidth]{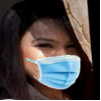}
\includegraphics[width=0.15\textwidth]{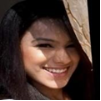}
\includegraphics[width=0.15\textwidth]{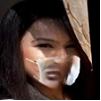}
\includegraphics[width=0.15\textwidth]{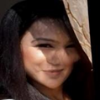}
\includegraphics[width=0.15\textwidth]{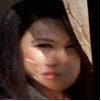}
\includegraphics[width=0.15\textwidth]{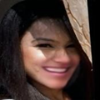}

\includegraphics[width=0.15\textwidth]{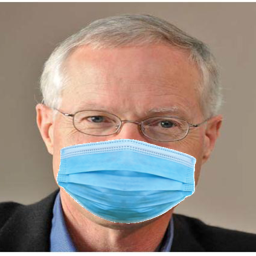}
\includegraphics[width=0.15\textwidth]{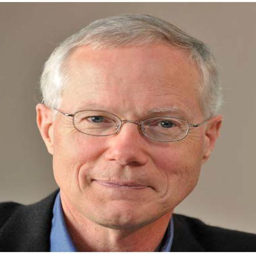}
\includegraphics[width=0.15\textwidth]{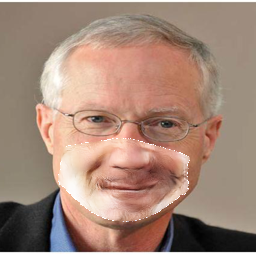}
\includegraphics[width=0.15\textwidth]{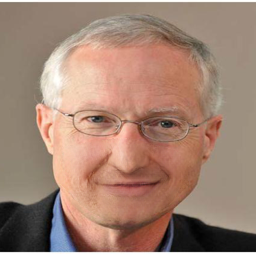}
\includegraphics[width=0.15\textwidth]{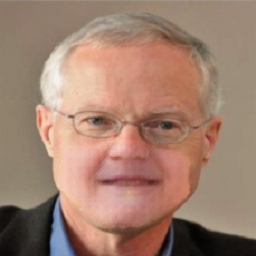}
\includegraphics[width=0.15\textwidth]{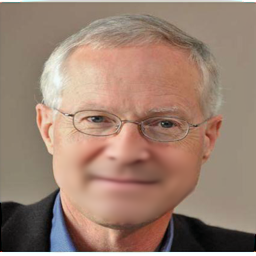}

\includegraphics[width=0.15\textwidth]{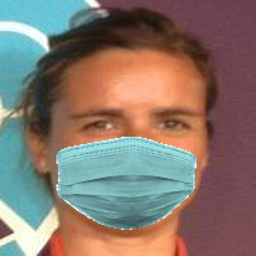}
\includegraphics[width=0.15\textwidth]{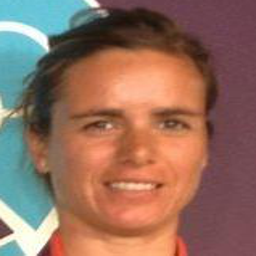}
\includegraphics[width=0.15\textwidth]{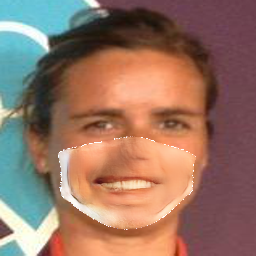}
\includegraphics[width=0.15\textwidth]{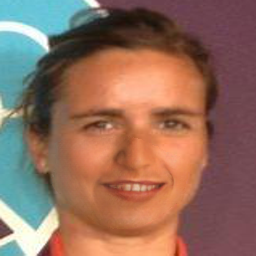}
\includegraphics[width=0.15\textwidth]{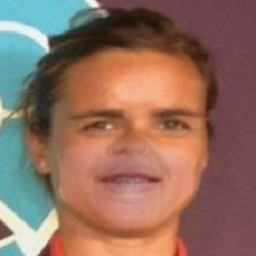}
\includegraphics[width=0.15\textwidth]{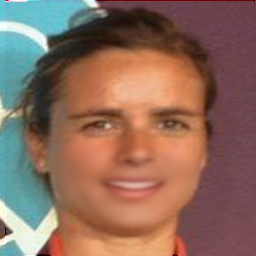}

\caption{\label{inpaintingQuality} \small Qualitative comparison of our results with state-of-the-art methods on CelebA dataset.}
\end{figure*}

\subsection{Attention Unit}
Attention is a method to highlight only the relevant activation's during training \cite{oktay2018attention}. It reduces the computational resources wasted on irrelevant activation and provides better network generalization. There are two types of attention: hard attention that highlights relevant regions and soft attention that emphasizes relevant and less on irrelevant areas. Soft attention is implemented at the skip connection shown in Fig. \ref{attention}. Attention gate takes 2 inputs: $x$ and $g$. Input $g$ is known as the gating signal, which is extracted from the previous layer with a lower dimension. Since $g$ comes from the middle layer, it has a better feature illustration.


On the other hand, $x$ comes from the skip connection. Since it comes from the early layer, it has better signal information. Aligned weights get larger after adding $x$ and $g$ while unaligned weights get relatively minor. Then, a ReLU activation function is performed followed by a $1\times1$ convolution operation denoted by $\psi$. The sigmoid activation is applied to scale the weights between 0 and 1 which are then up-sampled to maintain the original size of $x$. 

\subsection{Loss Function}

The performance of the deep learning model depends on the loss function. In our model, we have used SSIM loss with L1 normalization as loss function shown in Eq.\ref{loss}.
\begin{equation}\label{loss}
    ssim\_l1 =  SL + (L1*L1\_weight)
\end{equation}
where, SL indicates SSIM loss , L1 is mean squared error and L1\_weight stands for weight of L1 normalization.

\section{Experiment and Results}
\subsection{Dataset and Experimental Setup}
\textbf{Dataset.} To train and test our proposed model, we need large-scale face mask and unmask data from the same person. To address this issue, we use a public available CelebA \cite{liu2015faceattributes} dataset that contains 200k face-free data from celebrities around the world. We crop the face image at $256\times256$ size and create the synthetic mask \cite{anwar2020masked} on each face.

\begin{figure}[htb]
\centering
\includegraphics[width=0.115\textwidth]{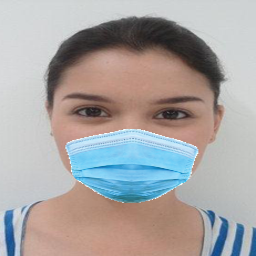}
\includegraphics[width=0.115\textwidth]{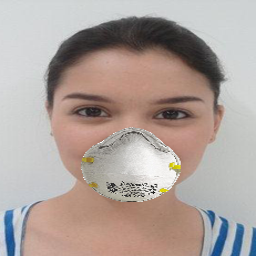}
\includegraphics[width=0.115\textwidth]{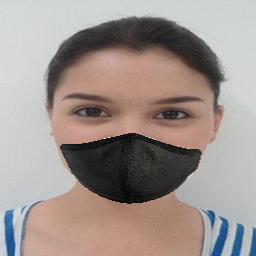}
\includegraphics[width=0.115\textwidth]{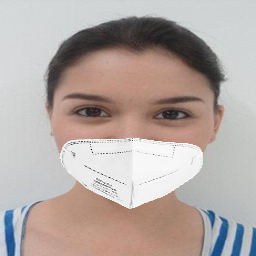}
\includegraphics[width=0.115\textwidth]{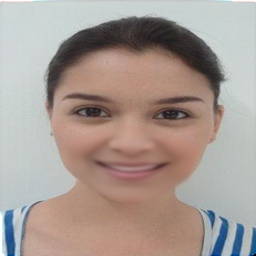}
\includegraphics[width=0.115\textwidth]{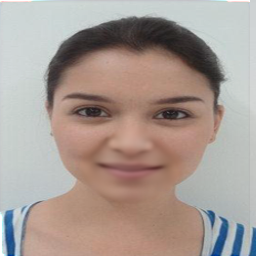}
\includegraphics[width=0.115\textwidth]{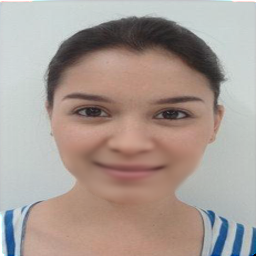}
\includegraphics[width=0.115\textwidth]{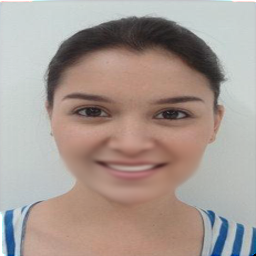}
\caption{\label{robustness} \small Robustness test with different types of mask. }
\end{figure}

\noindent \textbf{Experimental Setup.} The experiment was conducted using a Windows 10 workstation with 32 GB RAM and NVIDIA Geforce RTX 2070 GPU. We separate the testing data (10000 images) then the remaining data are divided 80 \% for training and 20\% for validation. We have run the experiment for 40 epochs with batch size 8. Adam optimizer and ssim\_l1 loss have been used as optimizer and loss function, respectively, with a learning rate of 0.0001.

\noindent \textbf{Evaluation Matrix.} Our method is evaluated by computing the Peak Signal to Noise Ratio (PSNR), Structural Similarity Index (SSIM), and Prediction Time. Higher PSNR and SSIM values indicate better quality in the output.

\subsection{Qualitative Evaluation}

Fig. \ref{Quality} shows the qualitative results of our face inpainting technique. Our method not only remove the mask completely but also reconstructs the faces with original structure.
In Fig.  \ref{inpaintingQuality}, we qualitatively compare our model with previous state-of-the-art techniques  \cite{yu2018generative,zheng2019pluralistic,din2020novel} for a set of test data. Yu et al. \cite{yu2018generative} fail to eradicate the face mask. On the other hand, Zheng et al. \cite{zheng2019pluralistic} improves the performance, but a noticeable face structural distortion occurs, e.g., images 1 and 2. The performance of Din et al. \cite{din2020novel} is acceptable. Unfortunately, it produces color inconsistency in the completion images. However, our model produces realistically pleasant outputs, maintains the original emotion such as sad and happy, and does not suffer noticeable color inconsistency. To demonstrate the robustness of the proposed method, we test it with different mask types. As shown in Fig. \ref{robustness}, our model produces realistic and similar image inpainting outputs on various masks. 

\begin{table}[htb]
\centering
\caption{Comparison of the quantitative results with state-of-the-art-methods on the CelebA dataset.}
\label{quantative}
\begin{tabular}{|c|c|c|c|}
\hline
\textbf{Method} & \textbf{SSIM}  & \textbf{PSNR} & \textbf{Prediction Time} \\ \hline

Yu et al. \cite{yu2018generative}  & 0.89  & 31.23 & 0.83 s /sample  \\ \hline
Zhen et al. \cite{zheng2019pluralistic}   & 0.92 & 33.31 & 1.15 s /sample     \\ \hline
Din et al. \cite{din2020novel} & 0.87 & 30.78 & 0.69 s /sample  \\ \hline

Ours  & \textbf{0.94} & \textbf{33.83} & \textbf{0.24 s} /sample  \\ \hline
\end{tabular}
\end{table}

\subsection{Quantitative Performance}
In TABLE \ref{quantative}, we report the quantitative evaluation results of our proposed method and state-of-the-art methods on the CelebA dataset. Our method has achieved the higher SSIM and PSNR at $0.94$ and $33.83$, respectively. The second-best performance received by Zhen et al. \cite{zheng2019pluralistic}. However, SSIM is more important than the PSNR for face inpainting tasks since it assesses the similarity between two images. For the Prediction Time, our model is about four times, five times, and three times faster than Yu et al. \cite{yu2018generative}, Zhen et al. \cite{zheng2019pluralistic}, and Din et al. \cite{din2020novel} respectively. 




\subsection{Ablation Study}
\textbf{Effect on Network depth.} More layers suffer higher parameters and training costs. For example, 3 layers with kernels $[32, 64, 128]$ have $0.58 M$ parameters and 4 layers with kernels $[64,128,256,512]$ have $9.68 M$ parameters. Furthermore, 6 layers with kernels $[32,64,128,256,512,1024 ]$ and $[64,128,256,512,1024,2048]$ have $39.14 M$ and $156.5 M$ parameters respectively. To better the performance and balance between performance and training cost, we investigated different architecture depths and received the best performance with 5 layers $[64,128,256,512,1024]$, $39 M$ parameters, and per epoch requires 530s.


\begin{table}[htb]
\centering
\caption{ \small Effect of Residual and Attention unit in the model.}
\label{TableAttention}
\begin{tabular}{|c|c|c|c|}
\hline
\textbf{Model} & \textbf{Parameters}  & \textbf{Loss} & \textbf{Accuracy}\\ \hline
UNet& 31 M & 0.07 & 0.90 \\ \hline
Attention UNet & 37 M  & 0.06 &  0.92 \\ \hline
RA UNet & 39 M  & 0.05 &  0.95\\ \hline
\end{tabular}
\end{table}

\noindent \textbf{UNet vs Attention UNet  vs  Residual Attention UNet.} We investigated the effectiveness of residual block and attention units in the model. We train on the same network to observe the efficacy except for the residual block and attention unit. TABLE \ref{TableAttention} shows how the performance varies with attention unit and residual block. Residual Attention UNet (RA UNet) achieved the best performance. 

\noindent \textbf{Choice of Loss Function.} Our method benefits significantly from the ssim\_l1 loss by its learning curve and faster convergence behaviors. Additionally, we have tested our model with Mean Square Error (MSE) and Mean Absolute Error (MAE) that took more than 3 times to converge compared to ssim\_l1 and received poor performance. 

\subsection{Failure Cases}
\begin{figure}[htb]
\centering
\begin{minipage}{1\textwidth}
\hspace{0.9cm} Input \hspace{2cm} GT \hspace{1.9cm} Output
\end{minipage}
\includegraphics[width=0.15\textwidth]{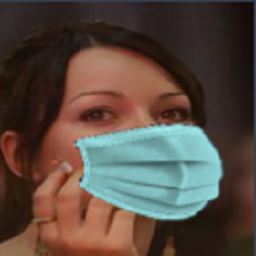}
\includegraphics[width=0.15\textwidth]{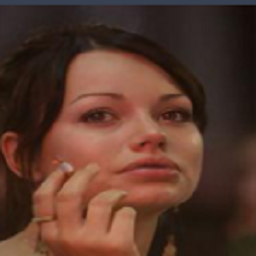}
\includegraphics[width=0.15\textwidth]{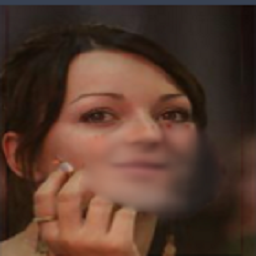}
\caption{\label{failCase} \small Example of a failure case.}
\end{figure}

However, our model can predict fast but not real-time. Besides, our method may fail to restore the face once the frontal view is unclear or occluded by other objects (e.g., hand pose). An example of the failure case has been shown in Fig. \ref{failCase}.

\section{Conclusion}
This paper presented a blind mask face inpainting method based on the residual attention UNet that automatically removes the mask and restores the face with fine details. We compared the proposed method with state-of-the-art methods both qualitatively and quantitatively, which shows the efficacy and robustness of our model. The limitation mentioned in the Failure Cases section will be considered in future works.

\section*{Acknowledgment}
This work is supported by the Scientific and Technological Research Council of Turkey (TUBITAK) under 2232 Outstanding Researchers program, Project No. 118C301.

\bibliographystyle{IEEEbib}
\bibliography{strings,refs}

\begin{thebibliography}{10}

\bibitem{deng2020image}
Ye~Deng and Jinjun Wang,
\newblock ``Image inpainting using parallel network,''
\newblock in {\em 2020 IEEE international conference on image processing
  (ICIP)}. IEEE, 2020, pp. 1088--1092.

\bibitem{yu2018generative}
Jiahui Yu, Zhe Lin, Jimei Yang, Xiaohui Shen, Xin Lu, and Thomas~S Huang,
\newblock ``Generative image inpainting with contextual attention,''
\newblock in {\em Proceedings of the IEEE conference on computer vision and
  pattern recognition}, 2018, pp. 5505--5514.

\bibitem{lahiri2019faster}
Avisek Lahiri, Arnav~Kumar Jain, Divyasri Nadendla, and Prabir~Kumar Biswas,
\newblock ``Faster unsupervised semantic inpainting: A gan based approach,''
\newblock in {\em 2019 IEEE International Conference on Image Processing
  (ICIP)}. IEEE, 2019, pp. 2706--2710.

\bibitem{zheng2019pluralistic}
Chuanxia Zheng, Tat-Jen Cham, and Jianfei Cai,
\newblock ``Pluralistic image completion,''
\newblock in {\em Proceedings of the IEEE/CVF Conference on Computer Vision and
  Pattern Recognition}, 2019, pp. 1438--1447.

\bibitem{li2020interactive}
Siyuan Li, Lu~Lu, Zhiqiang Zhang, Xin Cheng, Kepeng Xu, Wenxin Yu, Gang He,
  Jinjia Zhou, and Zhuo Yang,
\newblock ``Interactive separation network for image inpainting,''
\newblock in {\em 2020 IEEE International Conference on Image Processing
  (ICIP)}. IEEE, 2020, pp. 1008--1012.

\bibitem{din2020novel}
Nizam~Ud Din, Kamran Javed, Seho Bae, and Juneho Yi,
\newblock ``A novel gan-based network for unmasking of masked face,''
\newblock {\em IEEE Access}, vol. 8, pp. 44276--44287, 2020.

\bibitem{zhao2021comodgan}
Shengyu Zhao, Jonathan Cui, Yilun Sheng, Yue Dong, Xiao Liang, Eric~I Chang,
  and Yan Xu,
\newblock ``Large scale image completion via co-modulated generative
  adversarial networks,''
\newblock in {\em International Conference on Learning Representations (ICLR)},
  2021.

\bibitem{yang2020generative}
Yang Yang and Xiaojie Guo,
\newblock ``Generative landmark guided face inpainting,''
\newblock in {\em Chinese Conference on Pattern Recognition and Computer Vision
  (PRCV)}. Springer, 2020, pp. 14--26.

\bibitem{wang2020vcnet}
Yi~Wang, Ying-Cong Chen, Xin Tao, and Jiaya Jia,
\newblock ``Vcnet: A robust approach to blind image inpainting,''
\newblock in {\em European Conference on Computer Vision}. Springer, 2020, pp.
  752--768.

\bibitem{jiang2022mask}
Yefan Jiang, Fan Yang, Zhangxing Bian, Changsheng Lu, and Siyu Xia,
\newblock ``Mask removal: Face inpainting via attributes,''
\newblock {\em Multimedia Tools and Applications}, pp. 1--13, 2022.

\bibitem{liu2015faceattributes}
Ziwei Liu, Ping Luo, Xiaogang Wang, and Xiaoou Tang,
\newblock ``Deep learning face attributes in the wild,''
\newblock in {\em Proceedings of International Conference on Computer Vision
  (ICCV)}, December 2015.

\bibitem{ronneberger2015u}
Olaf Ronneberger, Philipp Fischer, and Thomas Brox,
\newblock ``U-net: Convolutional networks for biomedical image segmentation,''
\newblock in {\em International Conference on Medical image computing and
  computer-assisted intervention}. Springer, 2015, pp. 234--241.

\bibitem{he2016deep}
Kaiming He, Xiangyu Zhang, Shaoqing Ren, and Jian Sun,
\newblock ``Deep residual learning for image recognition,''
\newblock in {\em Proceedings of the IEEE conference on computer vision and
  pattern recognition}, 2016, pp. 770--778.

\bibitem{oktay2018attention}
Ozan Oktay, Jo~Schlemper, Loic~Le Folgoc, Matthew Lee, Mattias Heinrich,
  Kazunari Misawa, Kensaku Mori, Steven McDonagh, Nils~Y Hammerla, Bernhard
  Kainz, et~al.,
\newblock ``Attention u-net: Learning where to look for the pancreas,''
\newblock {\em arXiv preprint arXiv:1804.03999}, 2018.

\bibitem{anwar2020masked}
Aqeel Anwar and Arijit Raychowdhury,
\newblock ``Masked face recognition for secure authentication,''
\newblock {\em arXiv preprint arXiv:2008.11104}, 2020.

\end{thebibliography}

\end{document}